\documentclass[runningheads]{llncs}
\usepackage[T1]{fontenc}
\usepackage{graphicx}
\usepackage[acronym, symbols]{glossaries}
\usepackage{hyperref}
\usepackage{subcaption}
\usepackage{array}
\newcolumntype{P}[1]{>{\centering\arraybackslash}p{#1}}
\usepackage{booktabs}
\usepackage[justification=centering]{caption}
\usepackage{listings}
\usepackage{amsmath}
\usepackage{amssymb}
\usepackage{algorithm}
\usepackage{algpseudocode}
\usepackage{tablefootnote}

\makeglossaries
\newacronym{auc}{AUC}{Area under the receiver operating characteristic curve}
\newacronym{dgg}{DGG}{Discrete Global Grid}
\newacronym{kg}{KG}{Knowledge Graph}
\newacronym{osm}{OSM}{OpenStreetMap}
\newacronym{stkg}{STKG}{Spatial-Temporal Knowledge Graph}
\newacronym{id}{ID}{Identifier}
\newacronym{mcc}{MCC}{Matthews Correlation Coefficient}
\newacronym{de9im}{DE-9IM}{Dimensionally Extended 9-Intersection Model}
\newacronym{pca}{PCA}{Principal Component Analysis}
\newacronym{mae}{MAE}{Mean Absolute Error}
\newacronym{mape}{MAPE}{Mean Absolute Percentage Error}

\begin{document}
%
\title{GeoRDF2Vec -- Learning Location-Aware Entity Representations in Knowledge Graphs}
\titlerunning{GeoRDF2Vec -- Learning Location-Aware Entity Representations}
%
\author{Martin B\"ockling\inst{1}\orcidID{0000-0002-1143-4686} \and
Heiko Paulheim\inst{1}\orcidID{0000-0003-4386-8195} \and
Sarah Detzler\inst{2}\orcidID{0000-0002-7504-8856}}
\authorrunning{M. B\"ockling et al.}
%
\institute{Data and Web Science Group, University of Mannheim, Mannheim 68160, Germany \and
Corporate State University of Mannheim,  Mannheim, Germany}

\maketitle              
\begin{abstract}
Many knowledge graphs contain a substantial number of spatial entities, such as cities, buildings, and natural landmarks. For many of these entities, exact geometries are stored within the knowledge graphs. However, most existing approaches for learning entity representations do not take these geometries into account. In this paper, we introduce a variant of RDF2Vec that incorporates geometric information to learn location-aware embeddings of entities. Our approach expands different nodes by flooding the graph from geographic nodes, ensuring that each reachable node is considered. Based on the resulting flooded graph, we apply a modified version of RDF2Vec that biases graph walks using spatial weights. Through evaluations on multiple benchmark datasets, we demonstrate that our approach outperforms both non-location-aware RDF2Vec and GeoTransE.
\end{abstract}
\section{Introduction}
Knowledge Graph Embeddings are a family of methods designed to learn numeric, low-dimensional representations of entities in knowledge graphs. These representations can be used in downstream tasks to make predictions about entities, under the assumption that similar entities are mapped to similar embedding vectors~\cite{hubert2024similar,portisch2022knowledge}. Many large knowledge graphs not only encode relational information between entities but also capture the geographic geometries of some or all of these entities. Prominent knowledge graphs such as DBpedia~\cite{lehmann2015DBpedia}, YAGO~\cite{pellissier2020yago4}, and Wikidata~\cite{vrandevcic2014wikidata} contain geographic information on entities like places and buildings. Additionally, dedicated geographic knowledge graphs, such as KnowWhereGraph~\cite{janowicz_know_2022}, WorldKG~\cite{dsouza_worldkg_2021}, and OSMh3KG~\cite{bockling_planet_2024}, explicitly model geographic relationships.

When geographic relationships are explicitly represented—such as by assigning entities to grid cells and modeling spatial relations between these cells in the knowledge graph—they can be captured by embedding methods as well~\cite{bockling2024comparing}. However, most knowledge graphs store geographic information as literals using the WKT format, which is not well captured by most embedding methods~\cite{gesese2021survey}.

In this paper, we propose an approach that integrates geographic data into the well-known RDF2Vec method for knowledge graph embedding~\cite{Ristoski2016}. Our method consists of two steps. First, a geographic geometry is assigned to all entities in the graph, particularly those that either lack a predefined geometry or are not inherently geographic entities. Second, based on these geometries, the RDF2Vec graph walk mechanism is modified so that edges connecting geographically closer entities are assigned higher transition probabilities. We demonstrate that by incorporating geometric information, our approach achieves superior results compared to standard RDF2Vec while maintaining the same computational complexity as the original method.

\section{Related Work}
Within the domain of \glspl*{kg}, previous research has explored the use of spatial data in various tasks. Below, we provide an overview of different studies that have incorporated spatial data using various strategies for respective downstream tasks.

For geographical question answering, several models have been developed to generate spatially explicit \gls*{kg} embeddings. Unlike standard QA benchmark datasets, QA datasets that involve spatial data must account for additional constraints. One model that employs a specialized embedding approach for geographic data is TransGeo, an adaptation of TransE~\cite{Mai2019}. A similar approach is proposed by Qiu et al.\cite{Qiu2019}, who also present an adaptation of TransE. Like TransGeo, their method incorporates a penalization term using a spatial weight function. However, while Mai et al. use geodesic distance to compute spatial weights, Qiu et al. rely on Euclidean distance\cite{Qiu2019}. Since spherical distance functions provide more accurate distance calculations between geographic instances~\cite{ivis2006calculating}, we will focus on the TransGeo implementation proposed by Mai et al.

In TransGeo, the spatial weight is modeled using a stepwise function with a predefined distance threshold. If the distance exceeds this threshold, a constant value is assigned to the corresponding weight node instance. Geographic entities are identified by filtering DBpedia using the \textit{geo:geometry} property. Once the spatial weights for each edge are computed, PageRank is applied to the graph, using these edge weights to determine the PageRank value for each entity. This process captures the patterns of incoming and outgoing edges from specific entities. Finally, the computed entity weights are incorporated into an adapted objective function of TransGeo. Experimental results demonstrate that TransGeo outperforms various baselines in link prediction tasks as well as in QA datasets~\cite{Mai2019}.

Another approach, proposed by Mai et al.~\cite{Mai2020}, relies on an encoder-decoder architecture that captures various aspects of spatial information. By directly integrating spatial features—such as point coordinates and bounding boxes of geographic entities—into the embedding space, this method enables effective spatial reasoning.

The model consists of multiple components. An entity encoder separately captures both the semantics of the \gls*{kg} entity through type relations and its spatial embedding. If an entity has a point geometry, its coordinates are used as input for the spatial embedding. For non-point geometries, a point is randomly sampled from a uniform distribution within the bounding box. If an entity is not geographic, a random spatial embedding is assigned. These embeddings are then concatenated to form a unified entity embedding. Additionally, a projection operator maps entity-relation pairs to specific embeddings, enabling link prediction tasks. This allows the model to predict tail entities while also performing semantic reasoning to determine the most probable linked entity~\cite{Mai2020}.

Mann et al.~\cite{Mann2023} address the challenge of predicting spatial links in geographic \glspl*{kg}, which are often sparsely interlinked. Traditional link prediction methods typically rely on existing entity relations, which may be missing in such \glspl*{kg}. To overcome this limitation, the authors propose two approaches:

Supervised Spatial Link Prediction (SSLP): This method leverages spatial and semantic embeddings derived from the literal values of geographic entities to predict links.
Unsupervised Inductive Spatial Link Prediction (USLP): Unlike SSLP, this approach does not require labeled training data. Instead, it uses the haversine distance between geohash grid cell centroids to infer spatial links.
The authors evaluate these methods using WorldKG, demonstrating that both SSLP and USLP outperform existing state-of-the-art link prediction techniques. Their results highlight the effectiveness of incorporating spatial and semantic embeddings to enhance the completeness of geographic knowledge graphs~\cite{Mann2023}.

In this paper, we propose a location-aware variant of RDF2Vec\cite{paulheim2023embedding}. Our method modifies random walks by assigning different weights—and consequently, different transition probabilities—to edges. While prior studies have explored enhancing RDF2Vec with edge weights, using internal metrics such as PageRank\cite{cochez2017biased} or externally sourced relevance metrics~\cite{taweel2020towards}, our approach differs by leveraging geographic proximity as edge weights.


\section{Theoretical Background} \label{cha:TheoreticalBackground}

This section defines key concepts used throughout the paper and discusses various approaches for calculating weights between different spatial geometries.

\subsection{Definitions} \label{cha:Definitions}
In many public knowledge graphs, whether they model only spatial data or general information, spatial geometries are represented as literals within a \gls*{kg}. Our approach aims to leverage these geometries to model spatial relationships between individual nodes. To ensure consistency in terminology, we define key concepts in this section.
A knowledge graph $KG$ is defined as a directed labeled graph with a set of vertices $V$ and edges $E$. Therefore, it is defined as $KG=(V,E)$. The vertices $v$ are a finite set of individual nodes. Edges $E$ are defined as a finite set of individual edges connecting two nodes and having a relation type as a label, i.e., $E \subseteq V \times R \times V$, where $R$ is the set of different types of relations. 

In a $KG$, each node can has a \emph{neighborhood}. The neighborhood of a node $v \in V$ is defined as $H(v) := \left\{(s,p,o): \exists p,o | (v,p,o) \in E\right\} \cup \left\{(s,p,o): \exists s,p | (s,p,v) \in E\right\}$. The set of neighbors $N(v)$ of a node $v \in V$ is defined as \newline $N(v) := \left\{v' : \exists r | (v,r,v') \in H(v) \vee (v',r,v) \in H(v)\right\}$.

Geometries can be points, lines, polygons, multipoints, multilines, multipolygons and geometry collections. We define a \emph{spatial knowledge graph} $SKG$ as a knowledge graph where a part of the nodes has a geometry attached, i.e., $SKG = (V,E,S)$, where $S \subseteq V \times G$ is the set of nodes that have a geometry attached, and $G$ is the (theoretically infinite) set of all possible geometries. The subset $GV \subseteq V$ of geographic nodes is the set of nodes which have at least one geometry, i.e., $GV := \left\{v \in V | \exists g : (v,g)\in S\right\}$.

We further distinguish \emph{fully spatial KGs} as those KGs where \emph{every} node has a geometry attached (i.e., $V = GV$), and \emph{partially spatial KGs} where some, but not all nodes have a geometry attached ($GV \subset V$). The majority of common knowledge graphs, such as DBpedia or Wikidata, are partially geographic KGs.

Another important concept in our work are \emph{weighted knowledge graphs}. They are augmented with a weighting function $w : E \rightarrow \mathbb{R}^{+}$, which assigns a weight to each edge.

GeoSPARQL has been introduced as a standard for geographical representation in \glspl*{kg}. It defines six classes and 36 object properties. The superclass of GeoSPARQL is named \textit{SpatialObject}, which represents spatial entities and includes two primary subclasses: \textit{Feature} and \textit{Geometry}. The \textit{Geometry} subclass supports various geometry types and provides the ability to query them accordingly. Spatial geometries are represented as geographic literals, often using the WKT (Well-Known Text) format, which allows geometries to be encoded within the \gls*{kg}~\cite{Car2022}. In our approach, we adhere to the WKT standard for modeling geometries.
\subsection{From Spatial Distances to Edge Weights} \label{cha:SpatialWeights}

To calculate distances between two spatial entities \(i\) and \(j\), we use the \emph{geodesic distance}, which accounts for the ellipsoidal curvature of the Earth. Compared to other distance measurements, the geodesic distance provides a more accurate representation of the Earth's shape. In contrast, the \emph{great-circle distance} assumes the Earth is a perfect sphere. The difference between the great-circle distance and the geodesic distance becomes more pronounced over long distances, where the former introduces greater inaccuracies.

Therefore, we adopt the geodesic distance, as defined in \autoref{eq:geodesic_distance}. The distance \( d_{ij} \) between spatial entities \(i\) and \(j\) is computed using the Earth's radius \( R \) and the central angle \( \Delta \sigma \), as defined in \autoref{eq:delta_longitude}~\cite{Karney2012}. When using the WGS84 ellipsoid projection, the Earth's radius \( R \) is taken as a constant value of 6,378,137 m.
\begin{equation} \label{eq:geodesic_distance}
    d_{ij} = R \cdot \Delta \sigma
\end{equation}
To calculate the central angle between two points, the longitude \( \lambda \) and latitude \( \phi \) must be expressed in radians. The central angle \( \Delta \sigma \) represents the angular separation between two points on an ellipsoid, as observed from the ellipsoid's center. It is derived by combining latitude and longitude differences using trigonometric relationships.

The calculation of \( \Delta \sigma \) incorporates the difference in longitude \( \Delta \lambda \) and includes the sum of the product of the sine of each latitude \( \phi \), the cosine of each latitude, and the cosine of the difference between the longitudes \( \Delta \lambda \).
\begin{equation} \label{eq:delta_longitude}
    \Delta \sigma = \arccos (\sin \phi_{i} \cdot \sin \phi_{j} + \cos \phi_{i} \cdot \cos \phi_{j} \cdot \cos \Delta \lambda)
\end{equation}
Using these distances, we define a spatial weight matrix \( W \), where each entry \( w_{ij} \) represents the spatial weight between a spatial entity \( i \) and another spatial entity \( j \). One commonly used approach for spatial weighting is \emph{distance-based spatial weighting}. For a given distance function \( d_{ij} \) between spatial entities \( i \) and \( j \), the weight can be computed using various weighting functions. A widely used method is the \emph{threshold-based weight function}. Given a threshold \( \delta \), the pairwise function described in \autoref{eq:threshold_weight} is applied~\cite{Getis2009}.
\begin{equation} \label{eq:threshold_weight}
    w_{ij} = 
    \begin{cases}
        1 & \text{if } d_{ij} \le \delta \\
        0 & \text{if } d_{ij} > \delta
    \end{cases}
\end{equation}
Similarly to contiguity-based spatial weights, the threshold-based weight function assigns weights from the set \( \{0,1\} \). However, unlike these discrete spatial weighting methods, the \emph{inverse distance weighting} in \autoref{eq:inverse_weighting} and the \emph{exponential kernel function} in \autoref{eq:exponential_weight} calculate continuous weights. Inverse distance weighting employs a power parameter \( \alpha \in \mathbb{N} \), where a higher value of \( \alpha \) increases the influence of nearer points relative to more distant points. The value range of inverse distance weighting is within \( (0, \infty) \), where \( w_{ij} \in \mathbb{R}^{+} \)~\cite{Shepard1968}.

\begin{equation} \label{eq:inverse_weighting}
    w_{ij} = {(d_{ij})^{-\alpha}}
\end{equation}
In contrast to inverse distance weighting, the calculation of the spatial weight \( w_{ij} \) using an exponential function results in weights \( w_{ij} \) within the range \( (0,1] \). By using the inverse of the distance \( d_{ij} \), which falls within the value range \( (-\infty, 0] \), the exponential function assigns a value close to 0 for large distances and a value close to 1 for small distances.

\begin{equation} \label{eq:exponential_weight}
    w_{ij}=\exp(-d_{ij})
\end{equation}
When comparing different spatial weighting approaches, it becomes evident that each method offers specific advantages and disadvantages depending on the type of influence being measured. Inverse distance weighting has been shown to perform particularly well for global-scale measurements. In contrast, the threshold-based distance function (\autoref{eq:threshold_weight}) and contiguity-based weighting are limited to capturing only local dependencies, disregarding more distant influences. The exponential-based weighting, as described in \autoref{eq:exponential_weight}, provides a balance by allowing for \emph{quasi-globality}, where nearer spatial entities are assigned higher weights compared to those further away. Since our research focuses on interdependencies that are predominantly local due to the structure of the utilized \glspl*{kg}, we adopt the exponential function outlined in \autoref{eq:exponential_weight} to calculate spatial weights~\cite{Chen2012}.


\section{The GeoRDF2Vec Approach} \label{cha:Approach}
The GeoRDF2Vec approach consists of two major steps. First, to account for partially geographic knowledge graphs, we conduct a \emph{geographic information flooding} process to assign geometries to non-geographic entities. Based on these assigned geometries, we compute weights for all edges in the knowledge graph. Finally, we apply a weight-aware variant of RDF2Vec, where walks are extracted from the graph using the assigned weights to define transition probabilities.

\begin{figure}[t]
    \centering
    \includegraphics[width=0.8\linewidth]{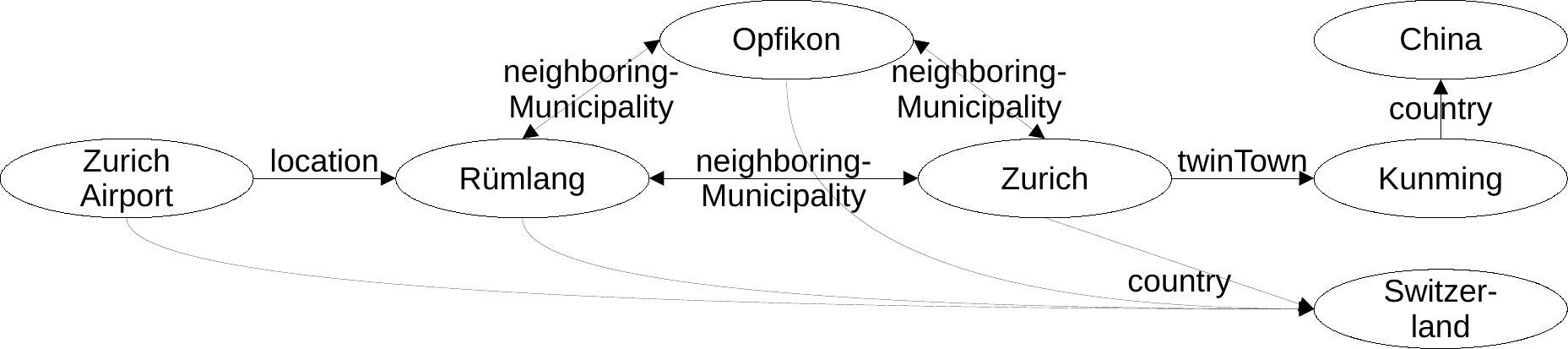}
    \caption{Example graph—excerpt from DBpedia}
    \label{fig:example_graph}
\end{figure}

The core idea behind GeoRDF2Vec is to assign higher weights to relations that connect geographically close entities. Considering the graph illustrated in \autoref{fig:example_graph}, classic RDF2Vec would extract the following walks with equal likelihood:
\begin{scriptsize}
\begin{verbatim}
Zurich Airport -> location -> Rümlang -> neighb.Mun. -> Opfikon -> neighb.Mun. -> Zurich
Zurich Airport -> location -> Rümlang -> neighb.Mun. -> Zurich -> twinTown -> Kunming
\end{verbatim}
\end{scriptsize}
However, the second walk is less useful for learning representations because Kunming is not a relevant entity for Zurich Airport, and vice versa. The main idea of GeoRDF2Vec is to reduce the likelihood of walks that connect geographically distant entities, as illustrated in the second example.

To visualize the concept of our approach, we use a generated Erd\H{o}s-R\'enyi graph with ten vertices, where two nodes are labeled as geographic nodes and are assigned random coordinates from Romania. In the following subsections, we provide an overview of how our approach is applied to the example graph.

\subsection{Geographic Information Flooding in Knowledge Graphs}\label{cha:GeographicFlooding}
Since most \glspl*{kg} are only partially spatial knowledge graphs, we first perform a preparatory step to assign geometries to non-geographic nodes. In this approach, each non-geographic node inherits the geometries of its geographic neighbors. Formally, the set \( S \) is extended as follows:
\[
S = S \cup \left\{(v,g) : v \notin GV \wedge v' \in n(v) \cap GV \wedge (v',g) \in S\right\}.
\]
Since some non-geographic nodes may still remain (i.e., those without a neighbor in \( GV \)), the process is iteratively repeated until no further geometries can be assigned.

We begin with the set \( GV \), which includes all nodes in the knowledge graph that already have a geometry. The direct neighbors of these nodes, which lack a geometry, are considered for processing in the next iteration. In each iteration, the set \( GV \) is expanded by assigning the geometries of neighboring nodes in \( GV \) to those without geometries. This process continues as long as there are non-geographic nodes adjacent to geographic nodes that have not yet been assigned a geometry.\footnote{Some nodes may lack a geometry at the end of the algorithm if their subgraph contains only non-geographic nodes, with no geographic neighbors to inherit from.} To propagate geographic information to all nodes in the network, we implement a flooding mechanism, as outlined in Algorithm~\ref{alg:GeoFlooding}. For efficient execution, the set \( S \) is stored in a hash table, using the node ID as the hash key for the corresponding values. As shown in \autoref{tab:GraphFlooding}, a single node can inherit multiple geometries if it has multiple geographic neighbors.

\begin{algorithm}[t]
\caption{Geographic information flooding}
\label{alg:GeoFlooding}    
\begin{algorithmic}
    \Require Partially geographic Knowledge Graph $(G,V,S)$
    \State $GV \gets \left\{v \in V | \exists g : (v,g)\in S\right\}$
    \Repeat
        \State $NodesToVisit \gets \left\{v \in V \setminus GV | \exists v' \in N(v) : v' \in GV\right\}$
        \For{$v \in NodesToVisit$}
            \For{$v' \in N(v)$}
                \For{$(v',g) \in S$}
                    \State $S = S \cup \left\{(v,g)\right\}$
                \EndFor
            \EndFor
        \EndFor
    \Until $NodesToVisit = \emptyset$
    \\\Return $(G,V,S)$
\end{algorithmic}
\end{algorithm}

To visualize the results, we start with our example where nodes (3, 9) are marked as geographic nodes. Applying the geographic information flooding algorithm to our generated sample graph, we observe that the algorithm terminates after two iterations, as all nodes have been assigned a geographic geometry. As shown in \autoref{tab:GraphFlooding}, once the flooding process is complete, three nodes (IDs: 4, 1, and 8) inherit both geometries. Nodes 5, 6, and 7 are assigned the geometry of node 3, while nodes 0 and 2 inherit the geometry of node 9.

\begin{table}[t]
    \caption{Iterations of the geographic flooding algorithm with the assignment of the geometries based on their node indices}
    \centering
    \begin{tabular}{P{0.3\textwidth}P{0.3\textwidth}P{0.3\textwidth}}
        \toprule
        \textbf{Iteration} & \textbf{Node IDs} & \textbf{Geometry Set} \\
        \midrule
        1 & 5,6,7 & \{3\} \\
        1 & 0,2 & \{9\} \\
        1 & 4 & \{9,3\} \\
        \midrule
        2 & 1,8 & \{9,3\} \\
        \bottomrule            
    \end{tabular}
    
    \label{tab:GraphFlooding}
\end{table}

\subsection{Spatial Weighting for Graph Walks} \label{cha:GraphSpatialWeight}
To assign weights to the edges, we apply the geodesic distance as defined in \autoref{eq:geodesic_distance}, in conjunction with the exponential weight function in \autoref{eq:exponential_weight}. As shown in \autoref{tab:GraphFlooding}, in certain cases, a node may be associated with multiple geometries. To account for this, we combine nodes with multiple geometries into a \emph{geometry collection}. This approach enables the integration of different geometry types into a unified set of geometries for each vertex \( v_{i} \) in a \gls*{kg} \( G \). Since the geodesic distance is defined between two points, we use the centroid of individual geometries or the centroid of the combined geometry set.

Due to the properties of the exponential function, large distances result in weights close to zero. For example, while a distance of 1 km corresponds to a weight of 0.3678, a distance of 5 km results in a weight of only 0.0067. In graphs with large spatial extents, many edge weights approach zero, limiting effective exploration. To mitigate this issue and enable graph traversal over larger distances, we normalize the distances.  For an individual node \( v_{i} \), the distances to all its neighbors are normalized using a \emph{min-max scaler}. Compared to other scaling methods, the min-max scaler preserves the distribution of vertex-neighbor distances. The min-max normalization of the distance \( d'(v_{i}, v_{j}) \) is defined in \autoref{eq:NormalizedDistance}.

\begin{equation}\label{eq:NormalizedDistance}
    d'(v_{i},v_{j}) = \frac{d(v_{i},v_{j}) -\min_{(s,p,o)\in H (v_{i})} d(s,o)}{\max_{(s,p,o)\in H (v_{i})} d(s,o) - \min_{(s,p,o)\in H (v_{i})} d(s,o)}
\end{equation}

\begin{figure}[t]
    \centering
    \begin{subfigure}{0.45\textwidth}
        \includegraphics[width=\textwidth]{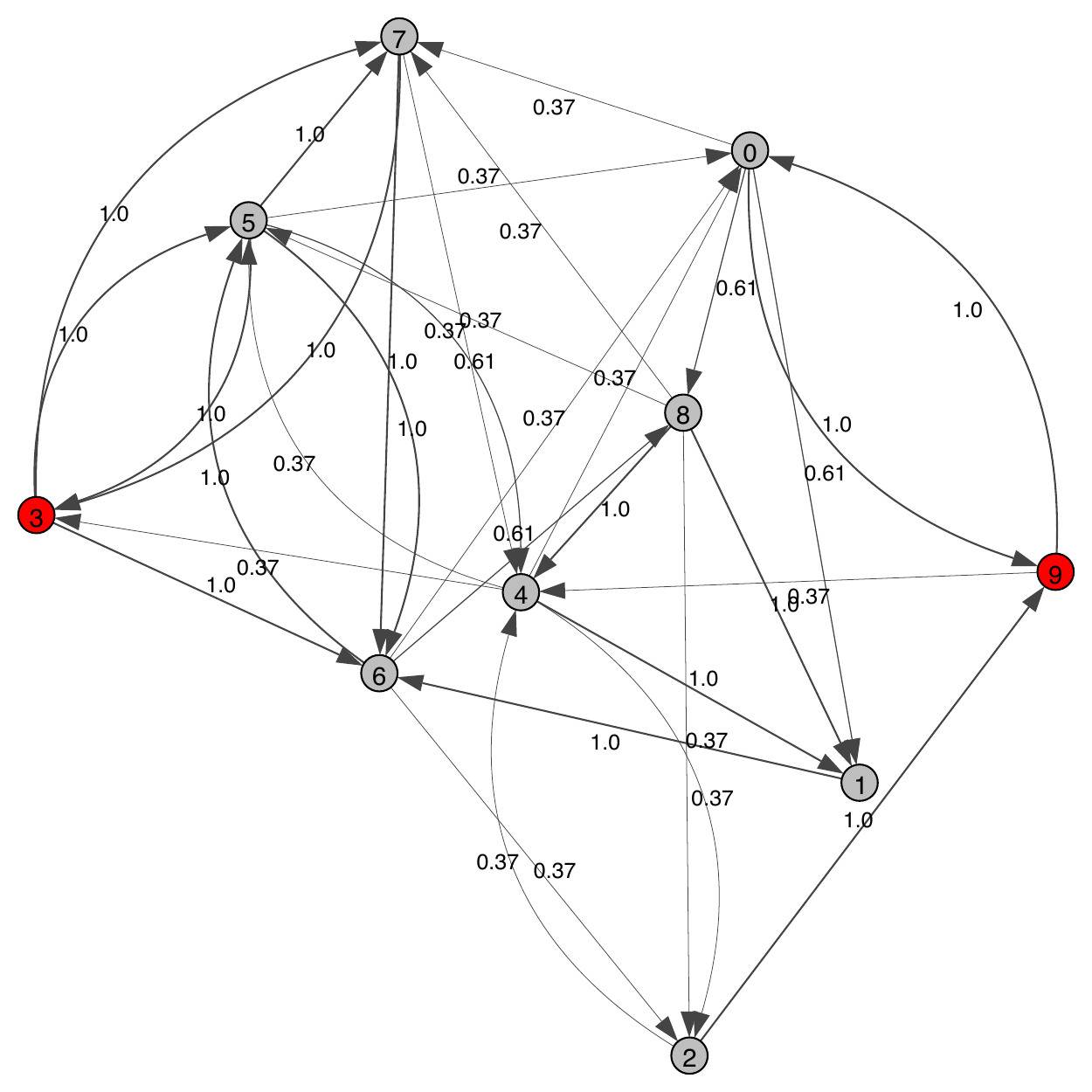}
        \caption{Normalized spatial weights for graph edges rounded to two decimal places} \label{fig:NormalizedWeightedGraph}
    \end{subfigure}
    \begin{subfigure}{0.45\textwidth}
        \includegraphics[width=\textwidth]{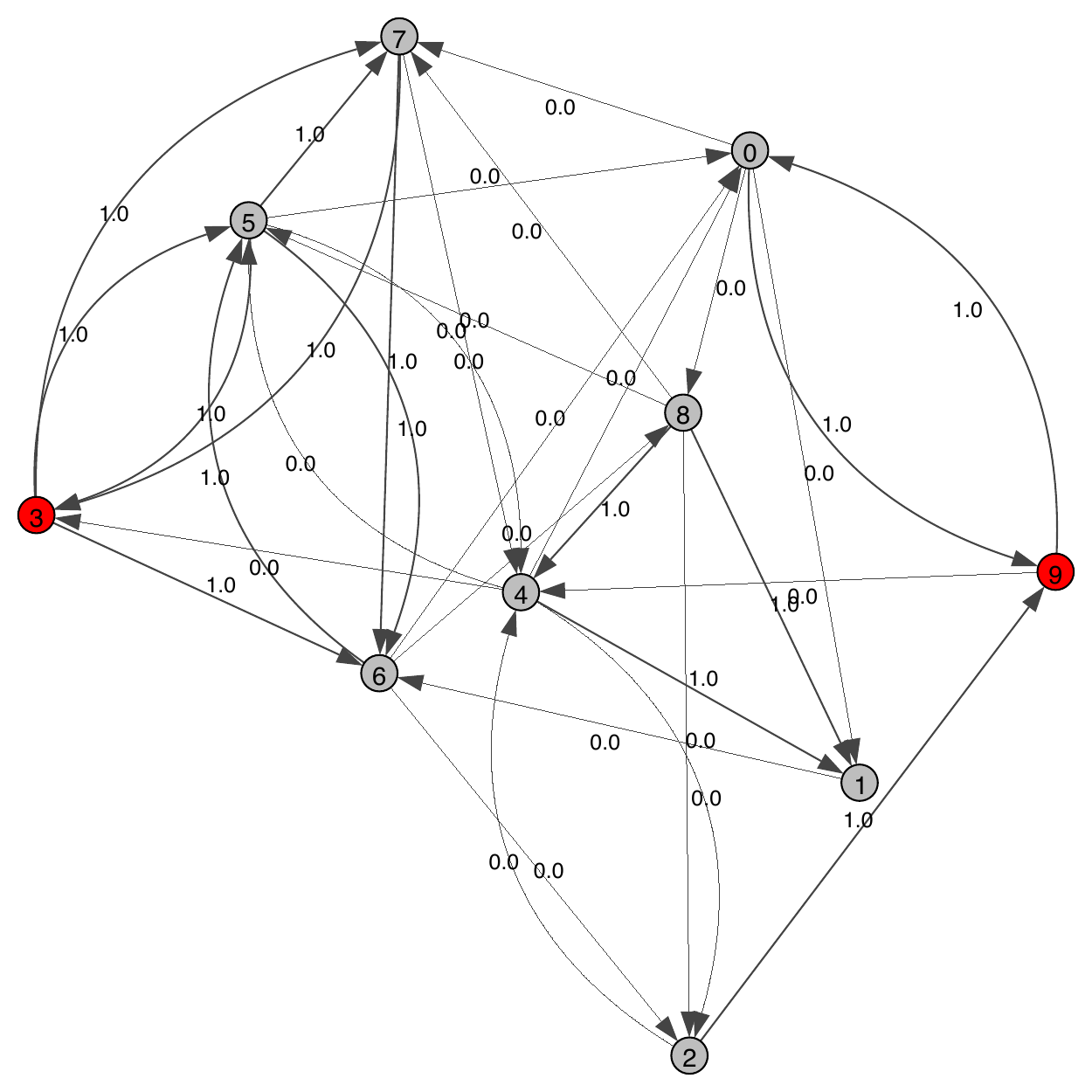}
        \caption{Non-Normalized spatial weights for graph edges rounded to two decimal places} \label{fig:UnnormalizedWeightedGraph}
    \end{subfigure}
    \caption{Different calculation of distances and resulting spatial weights for edges using the generated Erd\H{o}s-R\'enyi graph} \label{fig:WeightedGraph}
\end{figure}

The normalized distance \( d' \) is then used to compute the spatial weights. For each edge \( e_{i} \), we calculate the spatial weight \( w_{ij} \) and assign it to the corresponding edge \( e_{i} \). To visualize the effects of different weighting strategies on our constructed graph, we compute the spatial weight for each edge. In \autoref{fig:WeightedGraph}, the calculated weights are compared based on plain distance and normalized distance. Even after the flooding process, some nodes may remain without geographic coordinates. For edges connecting such nodes, a uniform weight of \( 1 \) is assigned.\footnote{This can only happen between two non-geographic nodes, not between a geographic and a non-geographic node, due to the flooding algorithm.}

\subsection{RDF2Vec on Weighted Knowledge Graph}\label{cha:WalkGeneration}
Based on the weighted graph, we apply RDF2Vec to generate an embedding for each individual vertex \( v_{i} \). RDF2Vec consists of two main steps: (i) the generation of graph walks and (ii) the computation of embedding vectors using word2vec. For the graph walks, two key parameters are considered. The depth \( d \), which defines the (maximum) length of each extracted walk and The number of walks per vertex \( w \). For the random walks, we use a transition probability matrix \( P \), where each entry \( p_{ij} \) represents the probability of transitioning from \( v_i \) to \( v_j \). It is defined in \autoref{eq:TransitionProbability}:

\begin{equation} \label{eq:TransitionProbability}
    p_{ij} = \begin{cases}
        \frac{w_{ij}}{\text{deg}(v_{i})} & \text{if} (v_{i},v_{j}) \in E, \\
        0 & \text{if} (v_{i},v_{j}) \not\in E.
    \end{cases}
\end{equation}
The weighted degree of a vertex $v_{i}$ is defined as $\text{deg}(v_{i})=\sum_{j=i}^{n}w_{ij}$. A single walk \( P_{v} \) is a sequence of nodes, where each subsequent node \( v_{j} \) is selected based on the transition probabilities between the connected nodes. The selection of neighboring nodes is repeated until either the random walk reaches a terminal node or the maximum walk distance \( d \) is reached. The complete corpus of extracted walks is the union of the maximum walks per entity and all extracted walks $\bigcup_{i=1}^{V}\bigcup_{i=1}^{w}P_{t_{i}}$. One single walk is represented by the set of combined triples $t_{i}$. \cite{Ristoski2016} The constructed walk corpus serves as input for the word2vec model. The word2vec model embeds each individual word in the corpus based on the generated walking sequences. Generally, word2vec differentiates between two training approaches: Skip-Gram and Continuous Bag of Words (CBOW). CBOW predicts a missing word from its surrounding words, while Skip-Gram predicts surrounding words given a specific word. \cite{Ristoski2016}

\section{Evaluation} \label{cha:Evaluation}

We evaluate our approach using two different benchmark frameworks. Additionally, we assess the effectiveness of the information flooding algorithm in isolation, as well as the impact of different hyperparameters.\footnote{The code and data for this paper are available \href{https://github.com/MartinBoeckling/GeoRDF2Vec}{here}.}

\subsection{Evaluation Frameworks} \label{cha:EvaluationFramework}

We utilize two evaluation frameworks: KGBench and GEval. KGBench provides a collection of curated \glspl*{kg} and evaluation tasks designed to assess relational or multimodal information encoding. For our study, we employ the dmg777k \gls*{kg} and its corresponding node classification task, as the DMG datasets in KGBench are the only ones containing spatial information. The dmg777k \gls*{kg} represents the locations of Dutch monuments using WKT geometries, which capture both the position and structure of each monument. The spatial extent of the \gls*{kg} is confined to the national boundaries of the Netherlands~\cite{Wilcke2020}. The node classification benchmarks for dmg777k include a total of five different node classes~\cite{Bloem2021}. To evaluate the classification performance on the dmg777k benchmark dataset, we use Accuracy, Macro F1 score, and the \gls*{mcc} score as evaluation metrics. As the classification algorithm for node classification, we employ XGBoost, optimizing its hyperparameters using Bayesian Search Optimization based on the recommended hyperparameters.\footnote{The selection of hyperparameters follows \href{https://xgboost.readthedocs.io/en/stable/tutorials/param_tuning.html}{XGBoost} recommendations.}

GEval is an evaluation framework designed to assess the performance of embedding approaches across various downstream datasets. It supports a range of learning tasks, including classification and unsupervised clustering. For each task, the framework specifies the evaluation metrics against which the results should be measured. GEval is built on a subset of DBpedia and assumes that the generated embeddings for entities are associated with DBpedia identifiers. The framework evaluates different embedding approaches using multiple algorithms, facilitating comparability across various methods~\cite{Pellegrino2019}. From the GEval evaluation framework, we utilize the three largest tasks: node classification, node regression, and node clustering.

For our evaluation of the TransGeo method, we implemented our own version based on the original paper, as the source code was not publicly available. The implemented code for TransGeo is provided within our GitHub repository. Following the original work by Mai et al., we heavily reused code from Wang et al., which implements a context-preserving variant of TransE~\cite{Wang2018}. The primary adaptation in our implementation was in the data preparation process, where we calculated entity-specific weights for context selection. For the experiments with TransGeo, we used the same set of hyperparameters as described in the original paper~\cite{Mai2019}.

\subsection{Knowledge Graph Characteristics}\label{cha:KGCharacteristics}
In the dmg777k \gls{kg}, two different geographic projections are used: the WGS 84 projection (EPSG:4326) and the Amersfoort / RD New projection (EPSG:28992). Among more than 20,000 geometries, 371 utilize the Amersfoort projection. Since these projections differ, we first convert the Amersfoort projection to the WGS 84 projection as a preprocessing step. Geographic entities using the Amersfoort projection are modeled as objects within a triple set with the specific predicate: \textit{http://data.pdok.nl/def/pdok\#asWKT-RD}. The dmg777k \gls*{kg} consists of 331,194 nodes and 776,920 edges. It is not a fully connected graph and contains multiple edges between certain vertices. The average degree of the graph is 4.6775. After performing geographic information flooding, the average geographic distance of an edge is 46.3675 km, and the average weight assigned to each individual edge is 0.8330~\cite{Bloem2021}.

For our experiments on GEval, we use the 2016 version of DBpedia, as GEval is based on this version. In DBpedia, geographic entities are implicitly modeled within the graph itself. In 2016, DBpedia used the GeoRSS standard for representing geographic data, which differs from the currently established GeoSPARQL standard that models geometries in the WKT format. To ensure compatibility, we transform all geometries present within the relevant subsets of DBpedia into the WKT format, allowing us to accurately identify geographic nodes within the \gls*{kg}. The DBpedia extract used in our experiments contains a total of 10,021,706 nodes and 84,087,224 triples. Similar to dmg777k, DBpedia is not fully connected and does not represent a simple graph. For our experiments with DBpedia, we utilize the following dataset components: Article categories, Instance types, Transitive instance types, Mapping-based object properties, SKOS categories.

\subsection{Evaluation Results} \label{cha:EvaluationResults}
For the dmg777k node classification task, we evaluate the multi-class classification performance using Accuracy, the Macro F1 score, and the MCC score. For GEval, we use the evaluation metrics assigned to each specific evaluation task.

\begin{table}[t]
    \centering
    \caption{Evaluation metrics for kgbench dmg777k node classification} \label{tab:dmg777k}
    \begin{tabular}{P{0.25\textwidth}P{0.25\textwidth}P{0.25\textwidth}P{0.25\textwidth}P{0.25\textwidth}}
        \toprule
        \textbf{Model} & \textbf{Accuracy} & \textbf{F1}  & \textbf{MCC} \\
        \midrule
        GeoRDF2Vec & \textbf{0.6672}$\pm 0.0203$ & \textbf{0.6060}$\pm 0.023$ & \textbf{0.5126}$\pm 0.0296$ \\
        RDF2Vec & $0.5882\pm0.0217$ & $0.536\pm 0.024$ & $0.4439 \pm 0.0282$ \\
        TransGeo & $0.5132\pm0.0212$ & $0.4901\pm 0.022$ & $0.4219 \pm 0.0280$ \\
        \bottomrule
    \end{tabular}
\end{table}

The results of our evaluation comparing different embedding methodologies on dmg777k are presented in \autoref{tab:dmg777k}. Across all evaluation metrics, GeoRDF2Vec outperforms both plain RDF2Vec and TransGeo, the latter serving as a spatially weighted alternative to our approach. 
Similar to other evaluation datasets, the translation-based method performs worse compared to RDF2Vec. While TransGeo implicitly models spatial weights associated with its nodes, its performance is inferior to the plain RDF2Vec implementation. This aligns with previous research, which indicates that non-simple \glspl*{kg} containing multiple predicates between the same set of entities tend to perform worse compared to RDF2Vec~\cite{Celebi2019}.  The spatially weighted variant of GeoRDF2Vec demonstrates significant performance improvements over plain RDF2Vec. Compared to TransGeo, our approach achieves a 26\% improvement in accuracy and a 21\% improvement in the F1 score.

\begin{table}[t]
    \centering
    \caption{Evaluation metrics for DBpedia GEval benchmark datasets} \label{tab:KGEvalResult}
    \begin{tabular}{P{0.18\textwidth}P{0.32\textwidth}P{0.22\textwidth}P{0.22\textwidth}}
        \toprule
        \textbf{Task} & \textbf{Dataset} & \textbf{RDF2Vec} & \textbf{GeoRDF2Vec}\\
        \midrule
        Classification & AAUP  & $0.676 \pm 0.0043$ & $\mathbf{0.775} \pm 0.0036$\\
          (accuracy) & Cities & $0.810 \pm 0.0128$ & $0.809\pm 0.0134$\\
         &  Forbes & $0.610 \pm 0.0093$ & $\mathbf{0.744} \pm 0.0152$ \\
         & Albums & $0.774 \pm 0.0053$ & $\mathbf{0.812} \pm 0.0049$ \\
         & Movies & $0.739 \pm 0.0037$ & $\mathbf{0.746} \pm 0.0029$ \\
        \midrule
        Clustering (accuracy) & Cities and countries (2k)  & $\mathbf{0.758} \pm 0.0038$ & $0.712 \pm 0.0029$\\
        & Cities and countries & $0.696\pm 0.0267$ & $\mathbf{0.812}\pm 0.0142$\\
         & Cities, albums, movies, AAUP, Forbes & $\mathbf{0.926}  \pm 0.0434$ & $0.821 \pm 0.0371$\\
         &  Teams& $0.917 \pm 0.0174$ & $\mathbf{0.928} \pm 0.0073$ \\
         \midrule
        Regression & AAUP  & $68.745 \pm 1.4563$ & $\mathbf{62.123} \pm 1.3591$ \\
         (RMSE) & Cities & $15.601 \pm 1.2734$ & $\mathbf{14.102} \pm 0.9571$ \\
         & Forbes & $36.459 \pm 1.0231$ & $\mathbf{32.241}\pm 1.2467$ \\
         & Albums & $\mathbf{11.930} \pm 0.0915$ & $12.131 \pm 0.0743$ \\
         & Movies & $19.648\pm 0.3428$ & $\mathbf{11.254}\pm 0.5310$ \\
        \bottomrule
    \end{tabular}
\end{table}

For the GEval benchmark, we evaluated all three approaches: GeoRDF2Vec, RDF2Vec, and TransGeo. Due to scalability limitations, TransGeo did not successfully scale to the full DBpedia \gls*{kg}. In its current implementation, the negative sampling process based on the entity context requires approximately 13 hours per sampling iteration on our hardware. A more scalable approach, such as using a subset of DBpedia or optimizing the negative sampling implementation, could improve feasibility for large \glspl*{kg} like DBpedia. However, such a reimplementation was beyond the scope of our research. As shown in \autoref{tab:KGEvalResult}, GeoRDF2Vec outperforms plain RDF2Vec in the majority of evaluated tasks.

\subsection{Analysis of the Information Flooding Approach}
For a deeper analysis of our information flooding approach, we tested five different queries on entities in DBpedia that do not have an associated geometry. These entities are linked to a country, for example, based on the location of an event or the nationality of a person. To ensure a manageable dataset size, we randomly limit the query results to 10,000 entities from each of the following queries:

\begin{scriptsize}
\begin{verbatim}
select ?x ?c ?g where {?x a dbo:Event . ?x dbo:country ?c . ?c geo:geometry ?g}
select ?x ?c ?g where {?x a dbo:Person . ?x dbo:nationality ?c . ?c geo:geometry ?g}
select ?x ?c ?g where {?x a dbo:RecordLabel . ?x dbo:location ?l . ?l dbo:country ?c . 
?c geo:geometry ?g}
select ?x ?c ?g where {?x a dbo:SportsTeam . ?x dbo:city ?y . ?y dbo:country ?c . 
?c geo:geometry ?g}
select ?x ?c ?g where {?x a dbo:Currency . ?x dbo:usingCountry ?c . ?c geo:geometry ?g}
\end{verbatim}
\end{scriptsize}

We determine whether the centroid of the geometry assigned to the entity \texttt{x}, computed as described in section~\ref{cha:Approach}, is in the geographic boundaries of the country \texttt{c}. The country boundaries are retrieved from OpenDataSoft \cite{opendatasoftWorldAdministrative}. If the centroid is within those boundaries, we count the example as a true prediction, otherwise as a false prediction. The resulting accuracies are shown in table~\ref{tab:flooding_accuracy}. We can see that in the majority of the cases, the information flooding assigns a sensible geometry to a non-geographic entity. The accuracy of the information flooding approach varies by query type, achieving 94.70\% for Event, 93.30\% for Person, 94.09\% for RecordLabel, 98.74\% for SportsTeam, and 84.86\% for Currency.


While the majority of cases in \autoref{tab:flooding_accuracy} achieve an accuracy above 90\%, we observe lower accuracy for the currency subset. This discrepancy arises due to the structure of the currency subset.
Incorrect geometries are assigned to currencies such as the New Zealand Pound, which has been used both in New Zealand and remote territories like Tokelau. Since DBpedia only contains point geometries, an unweighted mean is computed, distorting the centroid calculation. In contrast, for geometry collections that include polygons, the centroid would be weighted based on the area covered by each polygon.\footnote{The exact formula for centroid calculation can be found in the \href{https://shapely.readthedocs.io/en/2.0.6/reference/shapely.centroid.html}{documentation}.} To further evaluate the accuracy of our flooding approach, we conducted an additional experiment on the dmg777k dataset. We split the geographic nodes into an 80\% training set and a 20\% test set. Using the prior evaluation method, our flooding approach achieves an accuracy of 87.3590\% on the dmg777k dataset.

\subsection{Influence of GeoRDF2Vec Hyperparameters} \label{cha:GeoRDF2VecHPT}

\begin{figure}[t]
    \centering
    \begin{subfigure}{0.45\textwidth}
        \includegraphics[width=\textwidth]{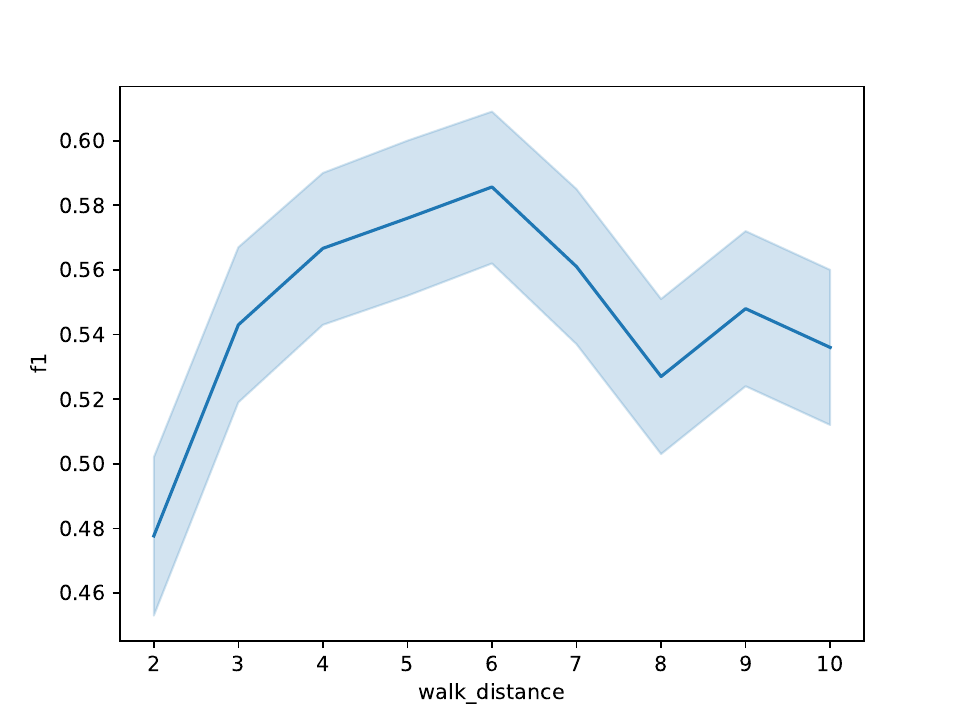}
        \caption{Maximum F1 score with CI for walk distance of RDF2Vec} \label{fig:FalseWalkDistance}
    \end{subfigure}
    \begin{subfigure}{0.45\textwidth}
        \includegraphics[width=\textwidth]{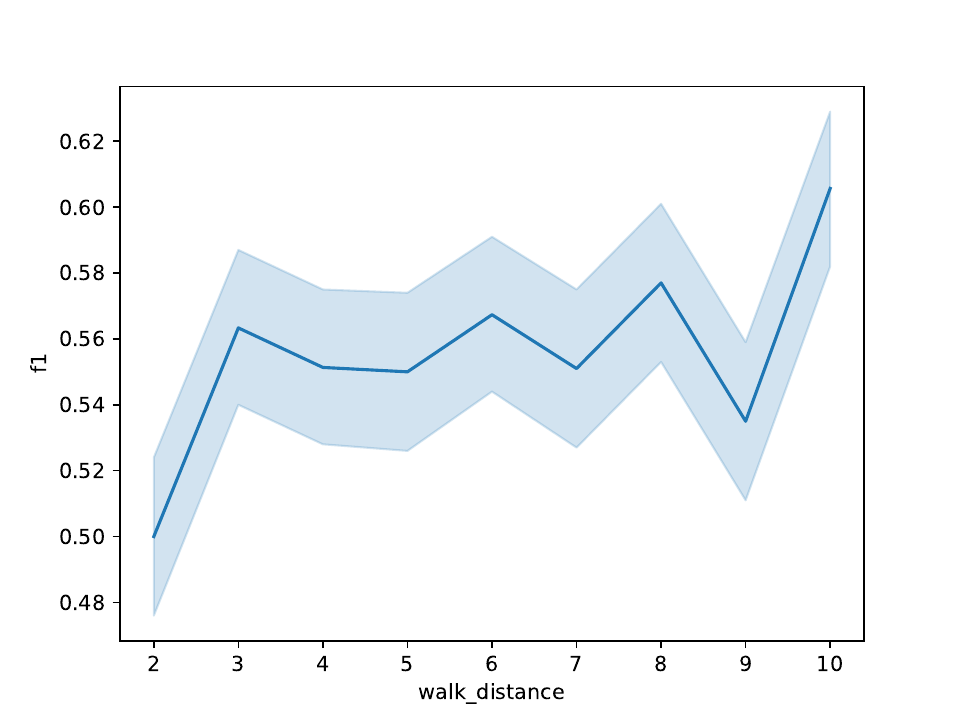}
        \caption{Maximum F1 score with CI for walk distance of GeoRDF2Vec} \label{fig:TrueWalkDistance}
    \end{subfigure}
    \begin{subfigure}{0.45\textwidth}
        \includegraphics[width=\textwidth]{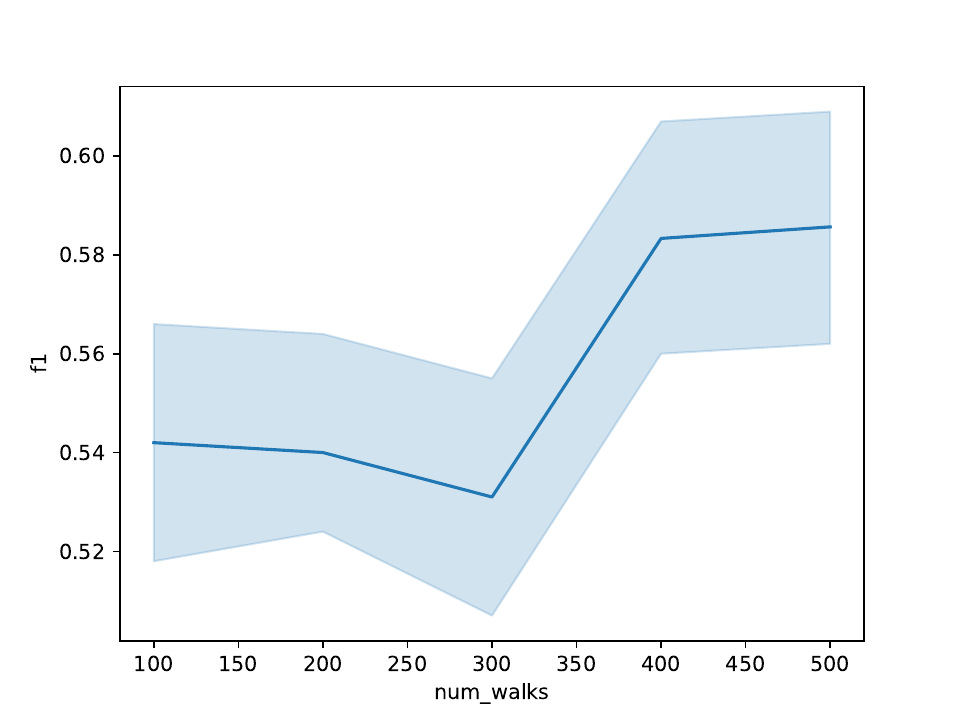}
        \caption{Maximum F1 score with CI for number of walks of RDF2Vec} \label{fig:FalseNumWalk}
    \end{subfigure}
    \begin{subfigure}{0.45\textwidth}
        \includegraphics[width=\textwidth]{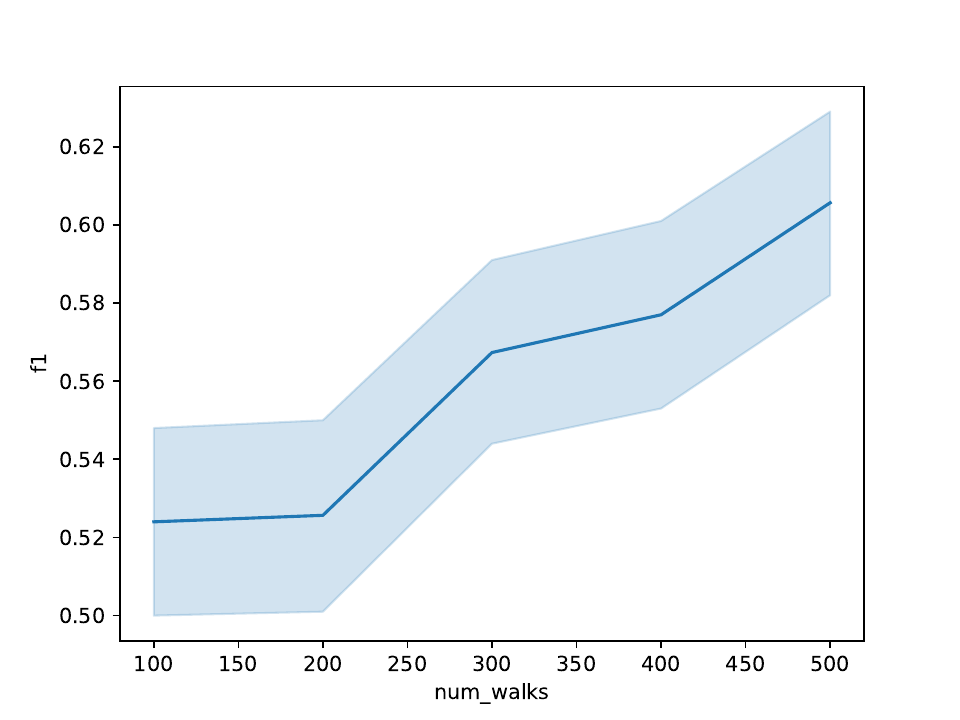}
        \caption{Maximum F1 score with CI for walk distance of GeoRDF2Vec} \label{fig:TrueNumWalk}
    \end{subfigure}
    \caption{Influence of walk distance and number of walks of RDF2Vec variants} \label{fig:WalkDistance}
\end{figure}

On the dmg777k node classification dataset, we conducted an ablation study to assess the influence of walk length and the number of walks in GeoRDF2Vec compared to the plain RDF2Vec implementation. To analyze the effect of both variables, we separate the grid search results for the weighted and unweighted graphs in the dmg777k evaluation dataset. We evaluate the grid search results by examining the maximum F1 score achieved for each individual parameter during the hyperparameter search. \autoref{fig:WalkDistance} illustrates the walk distance corresponding to the maximum F1 score for both the plain RDF2Vec variant and GeoRDF2Vec. For each parameter value, the confidence interval is visualized in the blue-shaded area of the line chart. 

For the plain RDF2Vec variant shown in \autoref{fig:FalseWalkDistance}, the F1 score increases from 0.48 to 0.58, reaching peak performance at a walk distance of 6. Beyond this distance, performance declines, with the F1 score decreasing to 0.54 at a walk distance of 10. In contrast, GeoRDF2Vec exhibits a similar initial performance increase, from 0.5 at a walk distance of 2 to 0.56 at a walk distance of 3. As depicted in the upper section of \autoref{fig:TrueWalkDistance}, the F1 score remains within the confidence interval until a walk distance of 7. However, at a walk distance of 10, the F1 score increases to 0.61.
These results indicate that introducing spatial weights during walk generation stabilizes performance and supports longer walk distances in graph embeddings. This allows for capturing more contextual information around individual entities, leading to more stable prediction results in downstream tasks. The likely reason for this improvement is that entities a few hops away in the graph can still be relevant if they are geographically close, whereas geographically distant entities are less relevant in the absence of spatial weighting.

The impact of the number of walks on RDF2Vec and GeoRDF2Vec is shown in \autoref{fig:WalkDistance}. In both variants, an increase in the number of walks leads to an improvement in the F1 score. However, the improvement is more substantial and continuous for GeoRDF2Vec compared to plain RDF2Vec. In standard RDF2Vec, all neighboring nodes are sampled with equal likelihood, leading to greater diversity in walk sequences as the number of walks increases. In contrast, when using geographically weighted walks, the diversity of walks is reduced since not all edges are sampled at the same rate. This less diverse sampling strategy, which prioritizes geographically proximal entities, evidently results in better performance.

\section{Conclusion and Outlook} \label{cha:Conclusion}
In this paper, we have demonstrated that incorporating geographic proximity when computing entity representations in knowledge graphs can significantly improve entity embeddings. By using geographic proximity as a proxy for transition probabilities in the walk generation mechanism, we have developed an effective, overhead-free method to integrate geographic distances into RDF2Vec. Additionally, through the geographic information flooding mechanism, we have introduced a way to leverage geographic information in partially geographic \glspl*{kg}. Previous research has proposed various walk generation mechanisms for RDF2Vec \cite{steenwinckel2021walk}. One such approach, \emph{community hops}, presents an interesting candidate for incorporating geographic information, as it could define hops based on geographic distance. While GeoRDF2Vec, as proposed here, only considers geographic proximity when entities share a direct connection in the knowledge graph, such an approach could also exploit geographic proximity for otherwise unconnected nodes.

Our approach incorporates spatial edge weights into the walk distance, filtering out entities that are geographically distant during walk generation. Future research could explore adapting established embedding methods such as TransE or ComplEx to incorporate geographic penalization in their loss functions. Specifically, introducing a penalization term for the geodesic distance between two entities could refine training processes by penalizing geographically distant entity pairs. Incorporating geographic penalization in loss functions has been shown to enhance learning outcomes when geographic data is integrated into a dataset \cite{peng2022domain}.
In addition to geographic information, many knowledge graphs also contain temporal facts, such as timestamps and timespans for events. In principle, our approach can be extended to temporal knowledge graphs as well.  Research has shown that, particularly in link prediction tasks, a deterministic approach using weights can outperform state-of-the-art models~\cite{Gastinger2024,krause2022generalized}. Ultimately, both geographic and temporal factors can be integrated into a unified weighting function to learn embeddings that incorporate both spatial and temporal aspects.

\section*{Resources}
The coding used for this paper together with datasets is provided via the following GitHub repository accessible under the following \href{https://github.com/MartinBoeckling/GeoRDF2Vec}{link}.



%
%
%
\bibliographystyle{splncs04}
\bibliography{bibliography}

\end{document}